\documentclass{article} 
\usepackage{iclr2017_workshop,times}
\usepackage{hyperref}
\usepackage{url}
\usepackage{graphicx}
\usepackage{amsmath}
\usepackage{caption}
\usepackage{subcaption}
\usepackage{tikz}
\usetikzlibrary{shapes.misc,arrows,positioning,shapes.geometric}

\title{Exploring loss function topology with \\cyclical learning rates}

\author{Leslie N.~Smith\\
	U.S. Naval Research Laboratory, Code 5514\\
	4555 Overlook Ave., SW., Washington, D.C.  20375\\
	{\tt\small leslie.smith@nrl.navy.mil}
	\And
	Nicholay Topin \\
	University of Maryland, Baltimore County \\
	Baltimore, MD 21250 \\
	\texttt{ntopin1@umbc.edu} \\
}

\begin{document}

	\maketitle
	
	\begin{abstract}
		We present observations and discussion of previously unreported phenomena discovered while training residual networks.
		The goal of this work is to better understand the nature of neural networks through the examination of these new empirical results.\footnote{Files to replicate these results are available at https://github.com/lnsmith54/exploring-loss} 
		These behaviors were identified through the application of Cyclical Learning Rates (CLR)  and linear network interpolation. 
		Among these behaviors are counterintuitive increases and decreases in training loss and instances of rapid training. 
		For example, we demonstrate how CLR can produce greater testing accuracy than traditional training despite using large learning rates. 
	\end{abstract}
	
	\section{Introduction}
	
	A core branch of physical sciences over the centuries has been the development of tools to experimentally probe invisible aspects of nature.
	Revolutionary discoveries, such as the structure of atoms and DNA, would not have been possible without clues from previous experimental evidence.
	
	Currently, most deep learning experimental results are reported in a limited number of standard ways.
	When using convolutional neural networks (CNNs) for classification, authors generally report the top-k accuracy or error/loss on held out test or validation samples after training the network.
	Optionally, a plot of these values over the course of training is provided.
	Though it is certainly important to report these results because they are a measure of success, we argue that it is valuable to take a new perspective: to investigate and report additional behaviors during training.
	
	\begin{figure} [tbh]
		\vspace{-5pt}
		\centering
		\begin{subfigure}[b]{0.47\textwidth}
			\includegraphics[width=\textwidth]{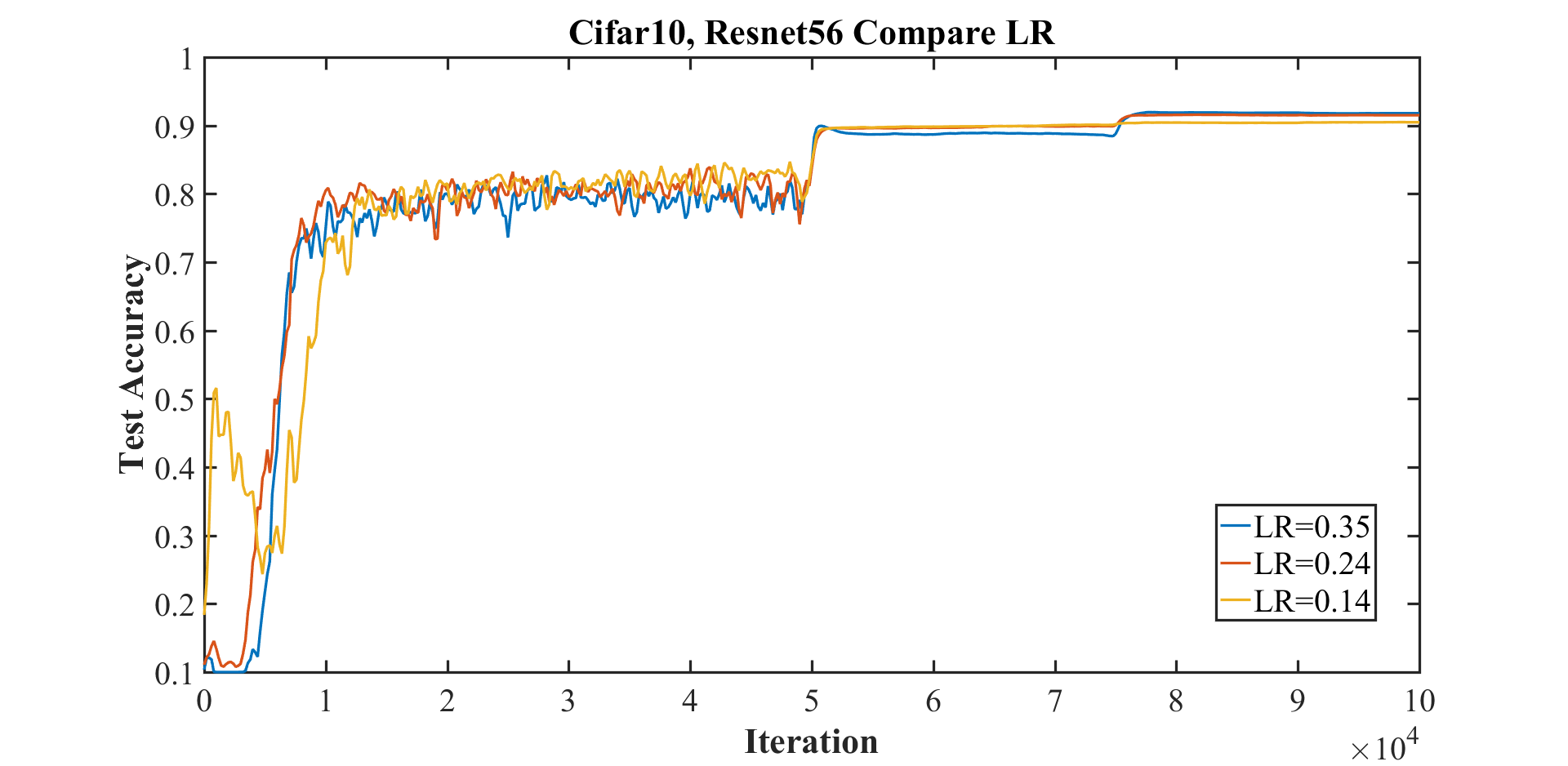}
			\caption{Test accuracy during standard training for three different initial learning rates.}
			\label{fig:Cifar10ResNet56AccLR}
		\end{subfigure}
		\quad
		\centering
		\begin{subfigure}[b]{0.47\textwidth}
			\includegraphics[width=\textwidth]{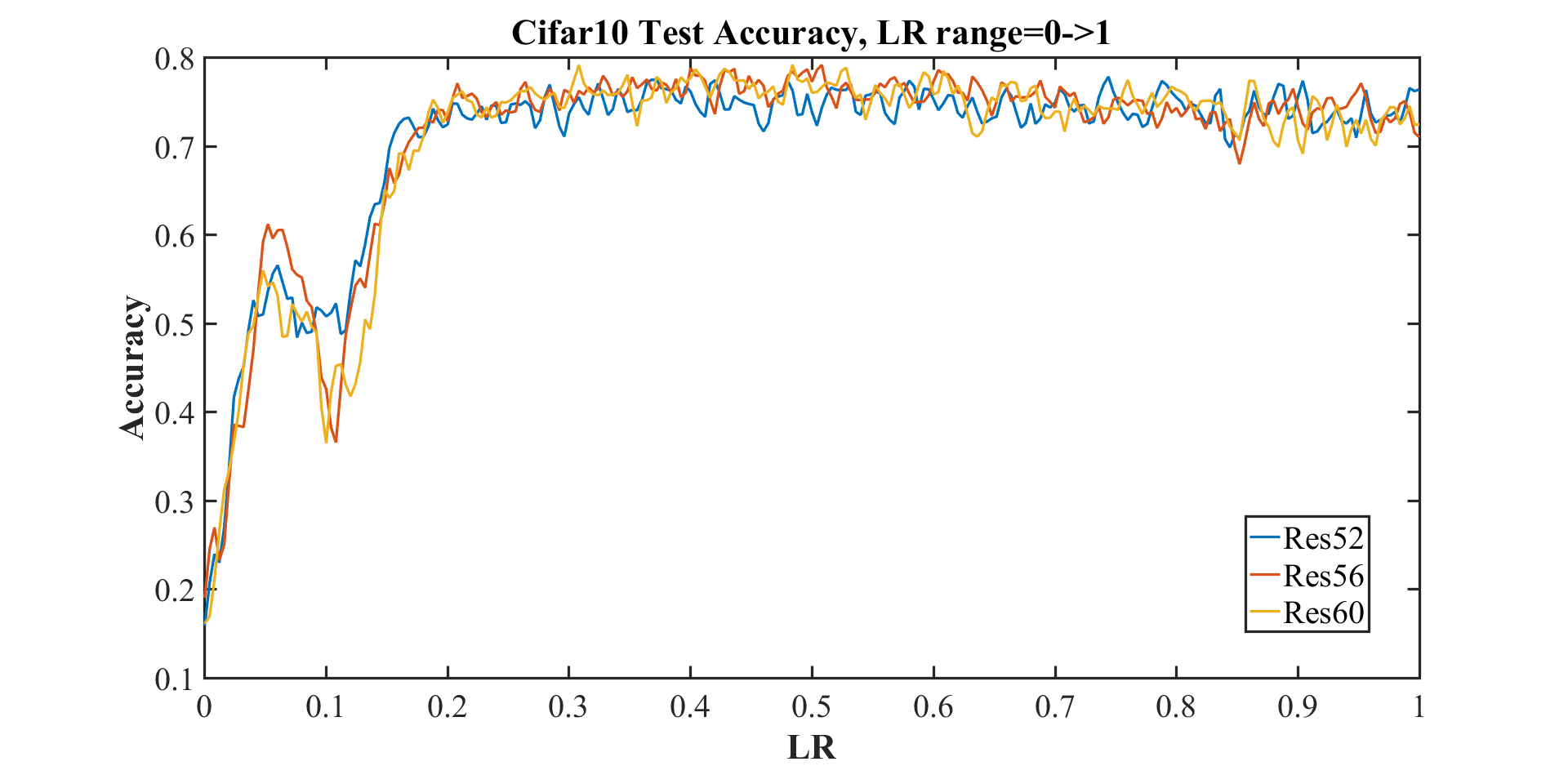}
			\caption{Test accuracy for learning rate range =0.001 - 1 for ResNet-52, ResNet-56, and ResNet-60.}
			\label{fig:ResNetCifar10Range1}
		\end{subfigure}
		\caption{Test accuracies for ResNet-56 on Cifar-10}
		\label{fig:Cifar10ResNet56_nonCLR}
	\end{figure}
	
	While running experiments on Cifar-10 with a ResNet-56 architecture using Caffe \citep{jia2014caffe} we noticed some unusual behavior and decided to investigate it.  
	Experiments with various learning rates (LR)  illustrate this unusual behavior, which can be seen in Figure \ref{fig:Cifar10ResNet56AccLR}.
	When using an initial LR of 0.14 (the yellow curve in Figure \ref{fig:Cifar10ResNet56AccLR}), the test accuracy climbs, then dips, and then continues to increase, which is unlike the curves for LR = 0.24 or 0.35.
	This strange phenomena caused us to look at the data in new ways and discover additional surprising results.
	
	The learning rate (LR) range test and cyclical learning rates (CLR) are described in \citep{smith2015cyclical} and  \citep{smith2017cyclical}.
	In a LR range test, training starts with a very small learning rate which is then linearly increased throughout training.  
	This provides information on how well the network converges over a range of learning rates.
	By starting with a small LR, the network starts converging, and as the LR rate becomes too large, it causes the training/test accuracies to decrease and the losses to increase.
	To investigate this dip in accuracy with certain learning rates, our first step was to run a learning rate range test (see Figure \ref{fig:ResNetCifar10Range1}), which displays two noteworthy features.
	First is the dip in accuracy around LR=0.1.
	Second is the consistently high test accuracy over a large span of learning rates (LR = $0.25$ to $1.0$), which is unusual.
	
	
	
	
	\begin{figure} [tbh]
		\vspace{-10pt}
		\begin{subfigure}[b]{0.47\textwidth}
			\includegraphics[width=\textwidth]{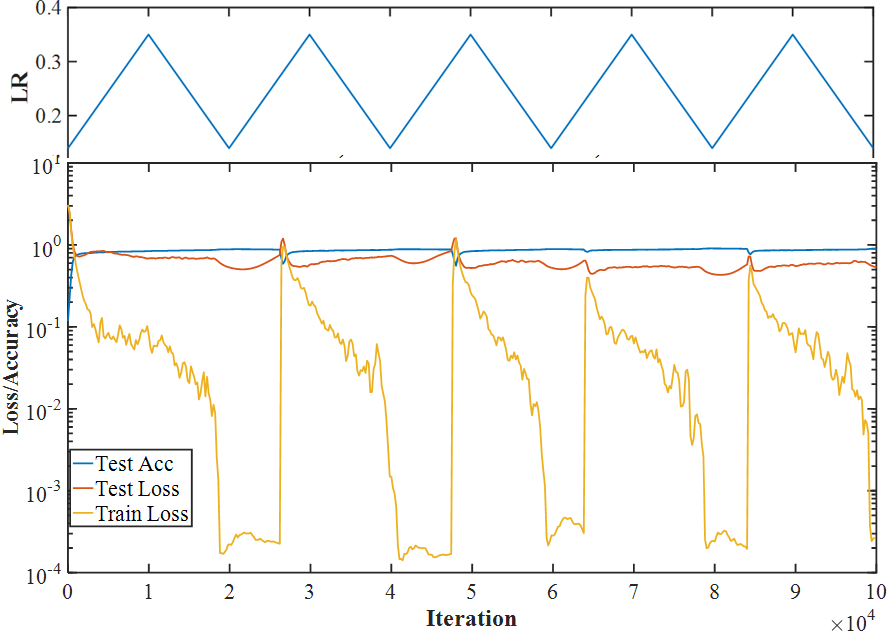}
			\caption{Cyclical learning rate between LR=0.1 and LR=0.35 with stepsize=10K.}
			\label{fig:Cifar10ResNet56TBS1000CLR35SS10k}
		\end{subfigure}
		\quad
		\centering
		\centering
		\begin{subfigure}[b]{0.47\textwidth}
			\includegraphics[width=\textwidth]{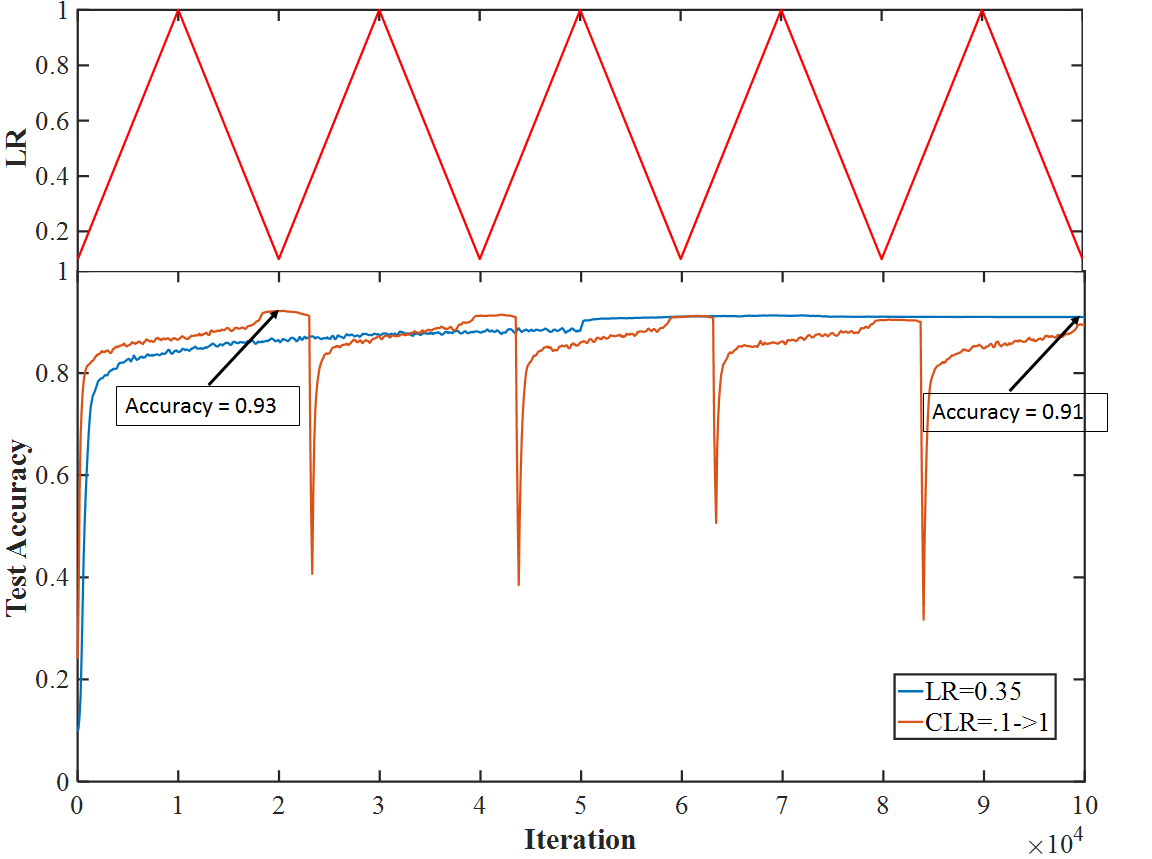}
			\caption{Super-convergence with CLR = 0.1 - 1.0 (stepsize=10k) versus standard training with initial LR=0.35.}
			\label{fig:Cifar10ResNet56TBS1000LR35CLR1}
		\end{subfigure}
		\caption{Test accuracies for ResNet-56 on Cifar-10.  Note the log scale used for the vertical axis.}
		\label{fig:Cifar10ResNet56_CLR}
		\vspace{-10pt}
	\end{figure}
	
	
	\section{Super-convergence}
	\label{sec:Super-convergence}
	
	Our next step was to try cyclical learning rates with a triangular policy for a number of cycles with the LR between 0.1 and 0.35.
	A triangular policy is a learning rate schedule that repeatedly linearly increases then decreases the learning rate between specified bounds (i.e.,  a minimum and maximum learning rate). 
	A cycle consists of two steps and the stepsize is the number of iterations over which the learning rate transitions from the minimum  to the maximum value or vice versa. 
	
	Several counterintuitive  results appear in Figure \ref{fig:Cifar10ResNet56TBS1000CLR35SS10k}, which shows the test accuracy, test loss, and training loss for CLR with a stepsize of 10,000 iterations.  
	This Figure shows an anomaly that occurs as the LR increases from 0.1 to 0.35.
	The training loss increases sharply by four orders of magnitude at a learning rate of approximately 0.255 (note the log scale used for the vertical axis) but training convergence resumes at larger learning rates.
	In addition, there are divergent behaviors between test accuracy and loss curves that are not easily explained.
	In the first cycle, when the learning rate is increasing from 0.13 to 0.18, the test loss increases but the test accuracy is also increasing.  
	This simultaneous increase in both the test loss and test accuracy also occurs in the second cycle as the learning rate decreases from 0.35 to 0.1 and in various portions of subsequent cycles.
	
	Another surprising result can be seen in Figure \ref{fig:Cifar10ResNet56TBS1000LR35CLR1}.
	The red curve in this figure shows a run with a triangular policy and a stepsize of 10,000 iterations (LR is between 0.1 and 1.0, as indicated at the top of the figure).
	The red curve shows rapid training of the network with a final test accuracy of 93\% in only one cycle of 20,000 iterations.
	For comparison, the blue curve shows a typical training process with an initial learning rate of 0.35, which drops by a factor of 10 at iterations 50,000, 75,000, and 85,000, and the final test accuracy is only 91\%.
	We coin the term ``super-convergence'' to refer to this phenomenon where a network is trained to a better final test accuracy compared to traditional training, but with fewer iterations and a much larger  learning rate.

	\section{Network Interpolation}
	\label{sec:network-interpolation}
	
\begin{figure} [tbh]
		\centering
		\begin{subfigure}[b]{0.47\textwidth}
			\includegraphics[width=\textwidth]{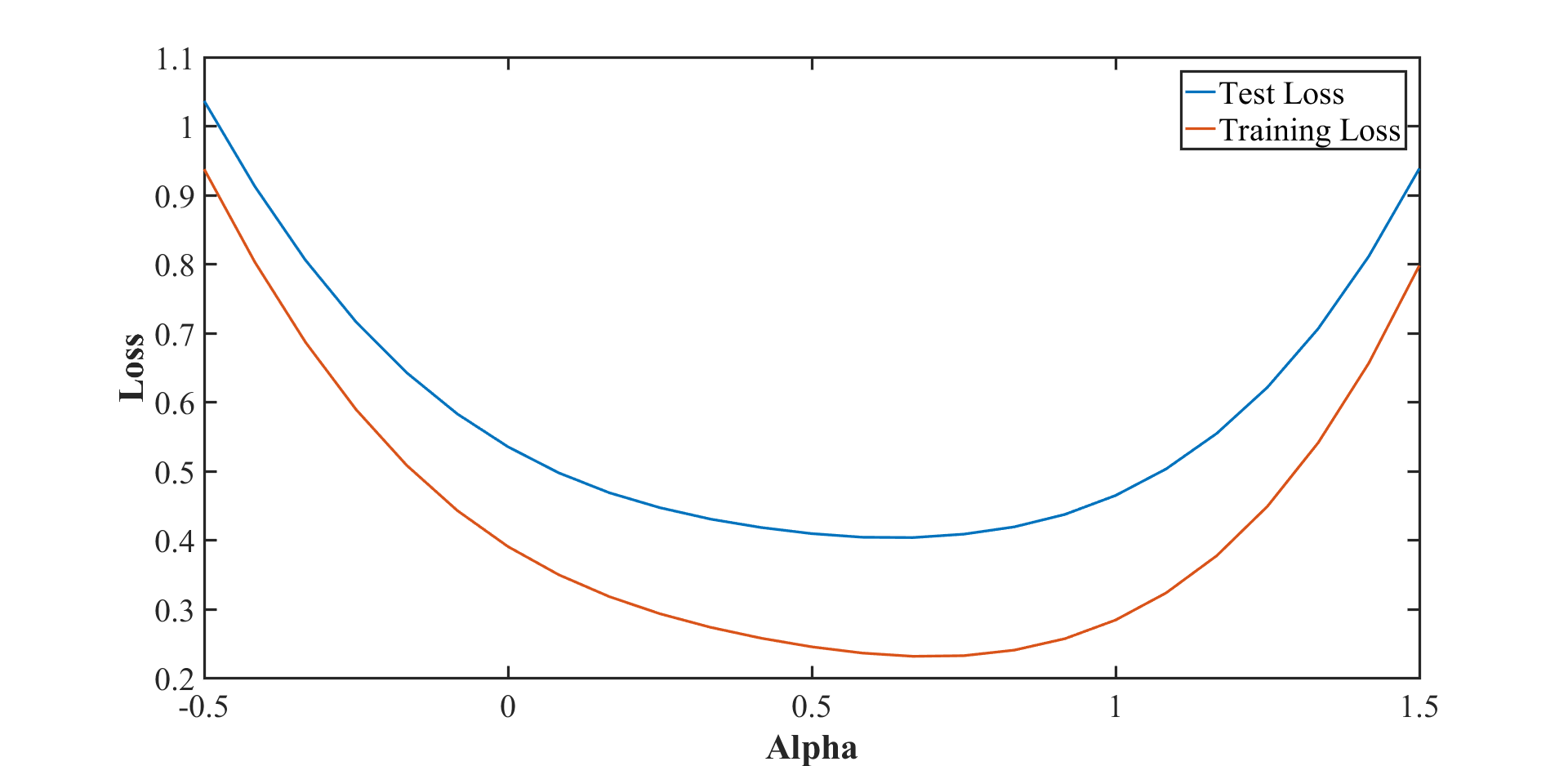}
			\caption{Training and test loss for interpolations between two network states from regular ResNet training after 10,000 and 25,000 iterations.}
			\label{fig:interpolation_regular}
		\end{subfigure}	
		\quad
		\centering
		\begin{subfigure}[b]{0.47\textwidth}
			\includegraphics[width=\textwidth]{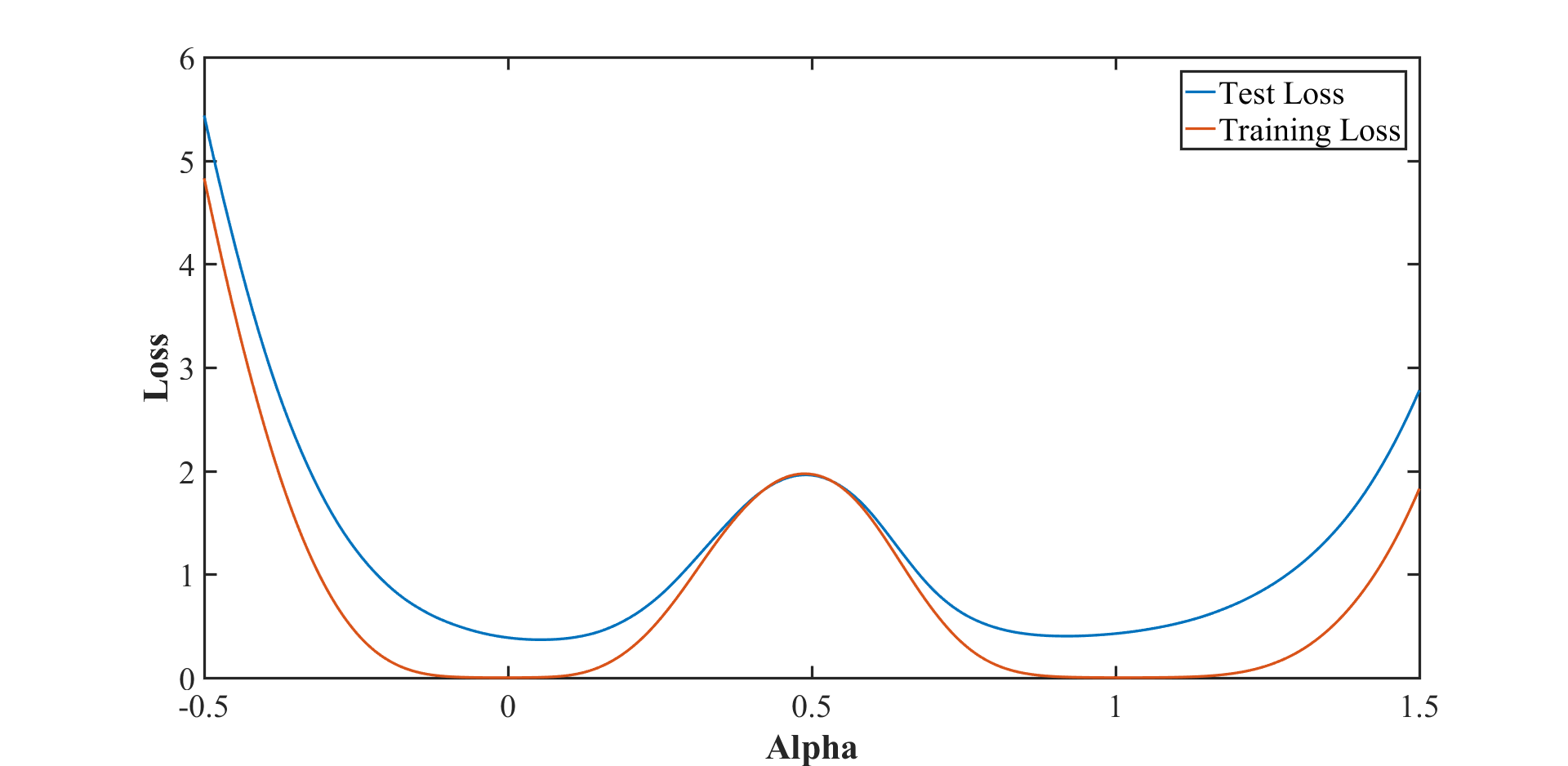}
			\caption{Training and test loss for interpolations between two CLR ``solutions'' - after 40,000 and 60,000 iterations, as seen in Figure \ref{fig:Cifar10ResNet56TBS1000LR35CLR1}.}
			\label{fig:interpolation_cyclical}
		\end{subfigure}
		\caption{Results of network interpolation for differently trained networks.}
		\label{fig:interpolation_together}
	\end{figure}

	On the basis of our findings described above, we believe it is reasonable to wonder if the solutions at each of the five peaks in  Figure \ref{fig:Cifar10ResNet56TBS1000LR35CLR1} are just the same minimum being rediscovered.
	We adopted the method of \cite{goodfellow2014qualitatively} and \cite{im2016empirical} to compare the solutions obtained at the end of each learning rate cycle (i.e., at iterations $20,000, 40,000, ...$).
	Briefly, a series of network configurations were created by performing an element-wise linear interpolation between a pair of solution weights (i.e., $\text{net}_{new} = \alpha*\text{net}_{1} + (1 - \alpha)*\text{net}_2$, for a range of $\alpha$ values).  
	If the pair of solutions represent the same minimum, interpolation should show a single concave shape, as seen in Figure \ref{fig:interpolation_regular}, which is the result of interpolating between states found during a regular training process. 
	If there is a ``peak'' in loss between the two solutions, as seen in Figure \ref{fig:interpolation_cyclical}, then the two minima are different solutions. 
	We found that the solution at the end of each cycle is distinct, which is in line with the results reported in \citep{im2016empirical}, where the authors show that different initializations lead to different solutions.
	Since the network is initialized differently at the beginning of each cycle, it is intuitive that each cycle produces a different solution. 
	
	The shape seen in Figure \ref{fig:interpolation_cyclical} reveals an additional  noteworthy feature: some amount of regularization is possible through interpolating between two solutions. The minima for \emph{training loss} are at $\alpha$ = 0.0 and 1.0, as expected, but the \emph{test loss} minima are slightly offset, towards the center. 
	This was consistent for all interpolations between minima found by CLR. 
	We are still investigating the causes of this phenomenon but it implies that interpolating between two such minima can improve performance and this might be useful for regularization. 
	
	Another relevant feature shown in Figure \ref{fig:interpolation_regular} is that the training and test loss minimum were consistently close to the center. 
	This is in line with the results shown in \cite{goodfellow2014qualitatively} (Figures 1 and 2), though the authors do not discuss this not its  potential to improve test accuracy. 
	This behavior leads us to believe that interpolation of weights at different iterations could be used to improve the quality of solutions found during regular training, in addition to those solutions found by CLR.




	


	\section{Conclusion}
	
	This paper shows  new phenomena obtained with ResNet-56 on Cifar-10 while using cyclical learning rates and the learning rate range test.
	We believe that the underlying reasons for these observed patterns are a reflection of the loss function topology and that a continuously changing learning rate provides information about this topology.
	Furthermore, we believe that this loss function topology information will lead to insights in training neural networks and we are actively searching for a collaborator to help us produce a theoretical analysis of these phenomena.
	
	The results presented here represent just a  fraction of the results we have obtained. 
	Similar results are obtained using ResNet-20, ResNet-56, and ResNet-110 on Cifar-10 and Cifar-100.
	We are cataloging the effect of different solvers, hyper-parameter values, and architectures.
	Every architecture, hyper-parameter value assignment, and data-set has its own patterns; some of them follow a usual pattern and others follow an unusual/unexpected pattern.
	In addition, we are investigating if the phenomena of high accuracies over a large range of learning rates (from the LR range test) might provide a measure of an architecture's robustness during training.
	While our work has so far been with Caffe, we plan to perform equivalent investigations with other frameworks (e.g., TensorFlow, Torch, MXNet).
	Furthermore, we are continuing to investigate how interpolation of network weights can be used to improve a network's performance. 
	The results of these investigations are being compiled into an NRL Technical Report and will be publicly released.

	
	\bibliography{exploringLoss.bib}
	\bibliographystyle{iclr2017_workshop}
	
\end{document}